# Challenges of Integrating A Priori Information Efficiently in the Discovery of Spatio-Temporal Objects in Large Databases

Benjamin Schott[1], Johannes Stegmaier[1],
Masanari Takamiya[2], Ralf Mikut[1]

[1] Karlsruhe Institute of Technology, Institute for Applied Computer Science,
[2] Karlsruhe Institute of Technology, Institute of Toxicology and Genetics
D-76344 Eggenstein-Leopoldshafen, Hermann-von-Helmholtz-Platz 1
E-Mail: benjamin.schott@kit.edu

## Motivation

Current research in areas such as life sciences, energy, developmental biology, traffic management and further highly actual fields often relies on the analysis of large amounts of spatio-temporal 2D+t and 3D+t data [1].

New developments in tracking technologies in combination with recent data acquisition techniques enable the collection of a tremendous amount of time resolved object localization data represented as trajectories in an unprecedentedly high resolution [2, 3]. Such collections of comprehensive object data enable the description of movement behavior and dynamics of the underlying objects. The traditional way of analyzing such object data is by visual inspection of human observers [4]. However, the tremendous amount of data yields to the problem that whole datasets can hardly be evaluated and analyzed manually [3]. Therefore sophisticated data analysis solutions are required (see therefore [5]). In significant applications in various domains (e.g. analyzing traffic datasets) such analysis solutions are required to extract information out of the collected data (e.g. identify and utilize groups of objects that exhibit similar patterns). The extracted information serves as a basis to gain knowledge about domain-specific object characteristics [6]. This is the basis for a deeper understanding of object behavior and underlying processes. To support this knowledge discovery process [7], different visual visualization can support the understanding of the analysis task and may serve as a feedback for the user [8]. There is a wide range of applications that require the detailed analysis of 2D+t and 3D+t object data. Due to advances in telecommunications, GPS technology, robotics, Web traffic monitoring and computer vision applications large amounts of spatio-temporal object data become more and more available [9]. The field of applications ranges from traffic management, surveillance and security [10, 11], airspace monitoring [12] to animal behavior, human movement [13], developmental biology [14] and social behavior studies [15] as well as sport scene analysis [6], to name only a few. The aim of such applications is, for example, the analysis of atmospheric phenomena and climate change, the optimization of transportation routes, a more detailed understanding of biological events in model organisms [14, 9], as well as the monitoring, modeling and simulation of dynamic phenomena [12, 16] and the improvement of tracking performance [17]. The large amounts of spatio-temporal data within the different types of applications introduce new challenges to data analysis and include the increasing need for novel approaches for the knowledge discovery process [18].

Regarding the analysis task, the degree of (dis)-similarity between trajectories is assessed using specific distance functions. On the basis of these trajectory similarities, often automated clustering methods are used to access the problem of gaining knowledge about the objects [19]. In the context of trajectory object data, clustering groups of trajectories based on their geometric proximity in spatial, temporal or spatio-temporal space. Here, the clustering can be based either on the raw trajectory data, extracted trajectory features or on models [20]. Examples of the different clustering techniques can be found in [19, 21, 22, 23]. The concept of multi-step clustering extends the simple one-layer clustering approach to multiple layers [24]. Here clustering can be applied in each layer to assess the characteristics of spatio-temporal objects more precisely. To cope with highly heterogeneous spatio-temporal objects, fully automated multi-step clustering approaches [25, 10] are not adequate enough. Therefore, interactive multi-step clustering approaches are introduced in [26, 27] to implement the application-specific *a priori* knowledge. In each step of such a cluster framework the user can choose different characteristics of objects to focus on. So it is possible to gradually build a comprehensive understanding of the set of complex objects [24, 26]. The current research therefore focuses on the combination of using human knowledge and apply-ing computer-based methods, such as clustering, to access the underlying information of objects.

However, the quantity and quality of *a priori* information varies according to the level of detail of the considered objects. Primarily on very high level of detail the quantity and quality of the *a priori* information is not sufficient enough to serve as a basis for semi-supervised methods such as clustering. Furthermore, depending on the focus of the analysis, a different validity of allocating information to the objects is required (e.g. only very rough trends in groups of objects on a lower level of detail are required, or on the contrary a very precise allocation of information on a high level of detail is needed).

To handle this problem we present a comprehensive framework starting with the raw object data over object detection and tracking up to the knowledge discovery process to extract valuable information. Due to the complexity of the heterogeneous behavior of objects, the human analyst can interact at various levels of detail in different ways (clustering, labeling, etc.) with the dataset to extract, analyze and visualize desired information directly from the interactively selected objects.

The remainder of this paper describes the framework in detail and specifies out the interaction interfaces. Next to the introduction of the framework, one possible application of interactive knowledge discovery is presented. The validation of the presented methods is performed using a benchmark dataset modelled on the basis of real biological data [31].

## Methods

In this section the general framework for designing the knowledge discovery process is described in general. Further the knowledge discovery process itself is explained more in detail.

### General Framework - Design the Knowledge Discovery-Process

A general scheme for designing the knowledge discovery (KD) process is shown in Figure 1. The aim is to generate a knowledge database containing all relevant information about the object database to answer a scientific question. To achieve this aim, the KD-process is designed iteratively to access the information within the object database. Due to the fact that *a priori* information often exists in different levels of detail within the object database, this stepwise approach leads to a very effective integration of the *a priori* knowledge. This is caused by the fact, that knowledge often appears during the KD-process on the basis of interim results. In contrary it would also be often possible to design the KD-process in only one step. However, creating such an one-step approach accessing all properties requires an enormous effort, irrespective of the interpretability. Further it is often not possible to integrate *a priori* knowledge efficient by such a one-step approach. Moreover, in each step of the framework, the problem formulation task serves as a basis for guiding the computational power towards the extraction of the required information. To support the interaction between user and framework, evaluation and visualization techniques serve as a feedback for optimizing the KD-process. In general the scheme can easily be extended and modified due to the modular architecture of the concept. The visualization techniques and the KD-process have further to be adapted to the actual application (e.g. to object classification and clustering in spatio-temporal datasets).

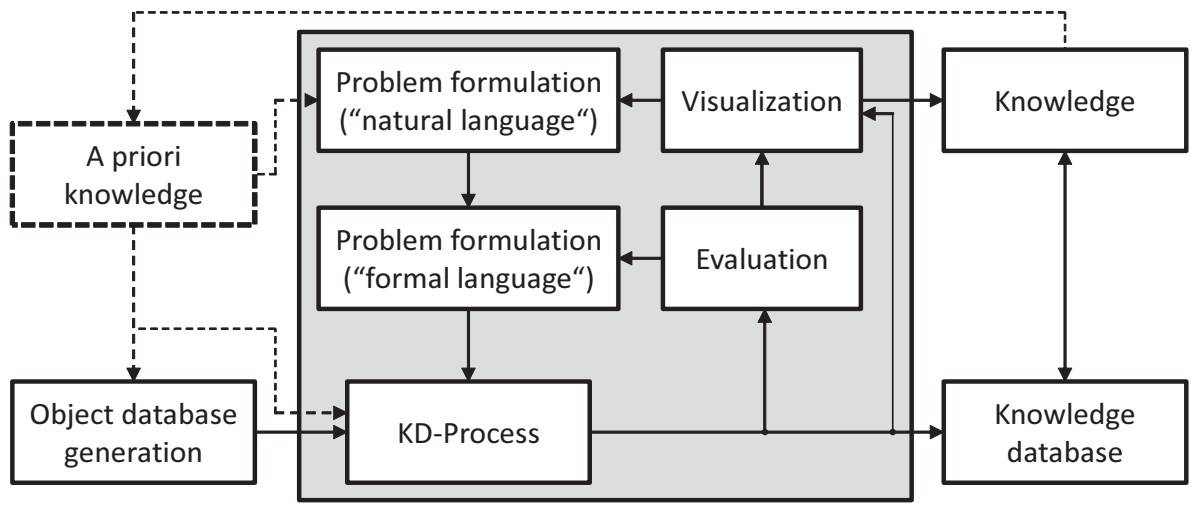

Figure 1: Scheme for the design of a knowledge discovery process. Here the effective integration of *a priori* knowledge to create a knowledge database is shown. Modified from [32].

Object Database Generation

In general an object database contains the data of several objects. In our case, the object database contains spatio-temporal object data. This serves as a basis for the KD-process for generating a knowledge database. There are many possible approaches to create such an object database, e.g. generate tracks of objects on the basis of raw object data, generate a benchmark on the basis of the raw object data or generate a data-independent benchmark (artificial benchmark, unrelated to any real world data), to name only a few. This processing step is modular and extensible, so further approaches to generate the object database can be added easily. In real world tracking projects however the tracks are not always complete over the full time period. This leads to artifacts based on short tracks. In this paper we assume a preprocessing to handle such problems.

Problem Formulation

Within each iteration of the KD-process, the first step is to specify the aims and restrictions, to identify an appropriate problem formulation in a natural language (e.g. find groups of objects that exhibit specific movement patterns). Based on this result, the natural language description of the problem has to be translated from the engineer into a more formal language (e.g. extract specific features of the objects), to be used within the algorithms of the framework. Such an interface, connecting both problem formulations, is fundamental to KD-process. This is an important prerequisite for an effective execution of the KD-process to extract meaningful knowledge out of the object databases.

Evaluation and Visualization

The evaluation and visualization of results in several processing steps serve as a feedback for the user. According to this feedback, the user is able to interact appropriately and to optimize the outcome of the process. The visualization is an important component because humans cannot deal well with large amounts of spatio-temporal object data. Therefore, visual rep-resentations are essential to support the human perception and reasoning process. This serves as a basis to assist the human analyst in understanding the object database.

To quantify the gathered results, the evaluation provides the human analyst with statistics of characteristics for groups of objects. Here, groups of objects can be evaluated according to selected spatial, temporal, spatio-temporal or object-based characteristics (e.g. color coding according features). This serves as a basis to compare several groups of objects and to derive statements about their behavior. This information can be used as additional feedback for the user to guide the KD-process.

**Knowledge Discovery Process**

The knowledge discovery process itself is shown in Figure 2. It consists of an information allocation process and the possibility for interactive modifications. For each step within the overall framework (Figure 1), there can be many iterations of information allocation and interactive modifications within the KD-process. Also the KD-process has a modular structure with standardized interfaces. This allows the modification and integration of methods within the process. Visualization and evaluation techniques serve again as a feedback for the user.

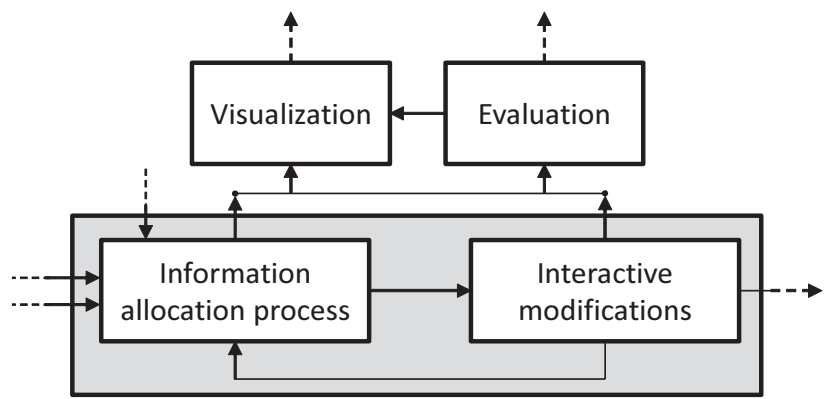

Figure 2: Scheme for the KD-process embedded within the general knowledge discovery framework in Figure 1.

## Information Allocation Process

In this processing step, information is assigned to every object within the database. This can be done by extracting characteristics for each object or by manually assigning specific information to each object. Based on the assigned information, the objects can be grouped according to similar characteristics of information. In general, there are different methods to allocate the information. In the following, three basic methods for information allocation (cluster analysis, interactive filtering and manual information allocation) are described more in detail.

In cluster analysis, objects are grouped according to predefined characteristics, either in spatial, spatio-temporal or other feature spaces, in such a way that objects belonging to the same group are more similar to each other compared to those objects belonging to other groups. Based on the extracted characteristics, the information is allocated to each object as the membership to a specific group. For the clustering method, there exist several approaches that can be chosen according to the underlying problem formulation, e.g. density-based clustering, hierarchical clustering, partitional clustering, fuzzy clustering as well as continuous online clustering approaches, to name only a few [19, 28, 29]. Furthermore all parameters of cluster analysis, like the similarity measure, the extracted characteristics and the cluster method itself are modular and can be combined within the framework. Additionally, new components can easily be added and implemented in the framework. This allows a very flexible and efficient investigation of the problem.

However, when *a priori* knowledge is vague and cannot be fully described through clear characteristics, an automated approach for information allocation, such as clustering, may be not appropriate anymore (e.g. there are no appropriate features available). This often happens on very detailed object levels suffering from low quality of the underlying *a priori* knowledge about expected or interesting results. In this case, the information can be directly assigned to the objects manually. In general, manual information allocation is an appropriate method applied to a detailed object level, containing only a few objects. To make the manual allocation process more effective, a visual framework is designed to support the user. Additionally, features or combinations of features that describe the manually labeled data the best can be extracted and used for further analysis.

The third method for information allocation is interactive filtering. In this approach parameters for filters can be set manually, including temporal filter (selection of a specific time interval), spatial filter (definition of a "window" in space), characteristic-based filter (selection of objects, based on their characteristics, such as length or curvature of the objects path) [24]. This allows a preselection of objects.

All three methods can be flexibly adapted and applied in different steps of the KD-process according the underlying problem formulation and the manifestation of *a priori* knowledge in quantity and quality.

Interactive Modifications

To optimize the allocated information in an iterative process, the possibility of interactive modification is required. If the results of the information allocation process are not satisfactory, the user has the ability to modify the results using his *a priori* knowledge. Mainly, the interactive modifications are used in the context of clustering. Here the user has the possibility to exclude objects belonging to one or several subclusters, fuse clusters or force a further clustering of selected clusters. Further options are to divide subclusters into smaller subclusters or to merge two or more subclusters to one larger subcluster. Moreover, there is the possibility to dissolve one or more subclusters and for example spread their members among the remaining subclusters [26]. The whole modification process has to be supported by appropriate visualization techniques and the possibility for appropriate interactions, e.g. focusing on subclusters or selecting one or more subclusters. Based on this, the user can optimally drive the KD-process to gain knowledge according the underlying problem formulation.

**Application of the Knowledge Discovery Process**

After the KD-process is designed it can be applied to new object databases. For a more detailed view of the process, see Figure 3. In each step, objects of the new database are classified according to the designed KD-process. Due to inhomogeneity and differences between the object databases, the user can, if necessary, modify the parameters within each step of the applied KD-process. This allows coping with small differences within various object databases.

To handle this problem, interactive modifications in each step can adapt the classification process. On a very high level of detail, objects from new object databases can often not be classified well, because the designed KD-process is based on another object database. To cope with this problem, the user can interact and add new information allocation steps. The result of applying the designed KD-process to new object databases is a new generated knowledge database.

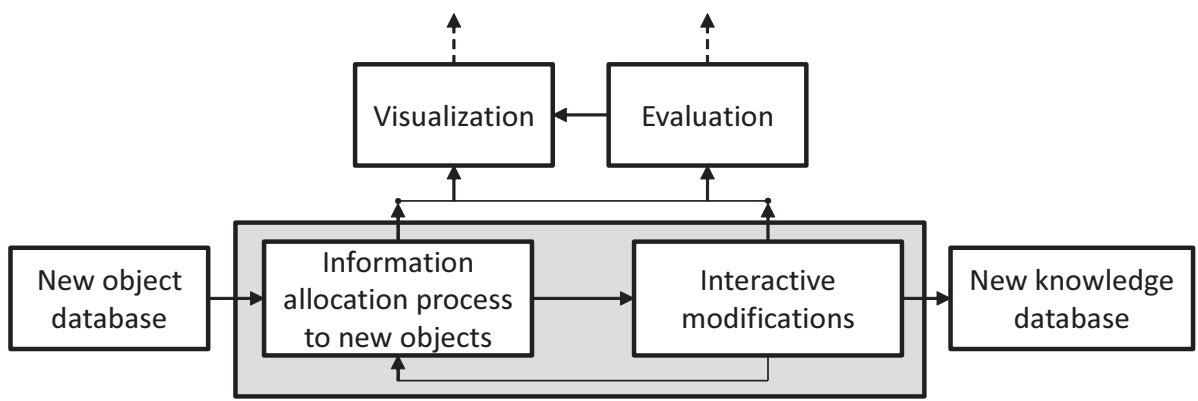

Figure 3: General scheme for the application of a KD-process. Here the process is displayed to create a new knowledge database out of a new generated object database

**Comparison of Different Object Databases**

For each object database a knowledge database is created using the KD-process. These knowledge databases can be compared and evaluated finally. Here quantitative statements about the comparison of the knowledge databases can be derived. This can serve as a basis for investigating for example different influences to object databases. Such influences are for example the application of different toxic compounds to biological organisms.

## Application

In the following, the knowledge discovery framework is exemplary applied to a benchmark object database to demonstrate its functionality. The overall problem formulation serves as a basis for generating the knowledge database. In some cases this is also possible using a one-step approach, instead of a multi-step framework. However, the approach would be much more complex and not interpretable any more. Further, the *a priori* knowledge within the human brain can hardly be transferred in the way that it can be implemented in such an one-step approach.
In the following sections, the stepwise integration of the *a priori* knowledge in the KD-process is explained in more detail (see Figure 4). Finally, the results of the KD-process and further analysis steps are described. All methods and analysis task used, are implemented using the Matlab toolbox Gait-CAD [3].

**Object Database Generation**

As an input for the knowledge discovery framework a benchmark object database is generated on the basis of object positions that were automatically extracted from real world biological raw object data [30]. Therefore, object data was extracted form light-sheet fluorescence microscopy images [3, 1]. For the benchmark generation, simulated objects (fluorescently labeled cell nuclei) are placed at randomly selected locations of real objects and temporal dynamics of the objects are simulated. For more details see the description in [31]. The advantage of such a semi-synthetic benchmark is the direct availability of on error-free ground truth combined with realistic object movement behavior. Additionally, the benchmark simulation framework allows the generation of artificially flawed data and different scenarios [31]. In Figure 5, the generated benchmark object database is shown. Here, 520 objects can be observed that exhibit complex movement patterns. Each object has 400 time points and the time series database has a size of approximately 8 MB and the related 3D image database has a size of approximately 250 GB.

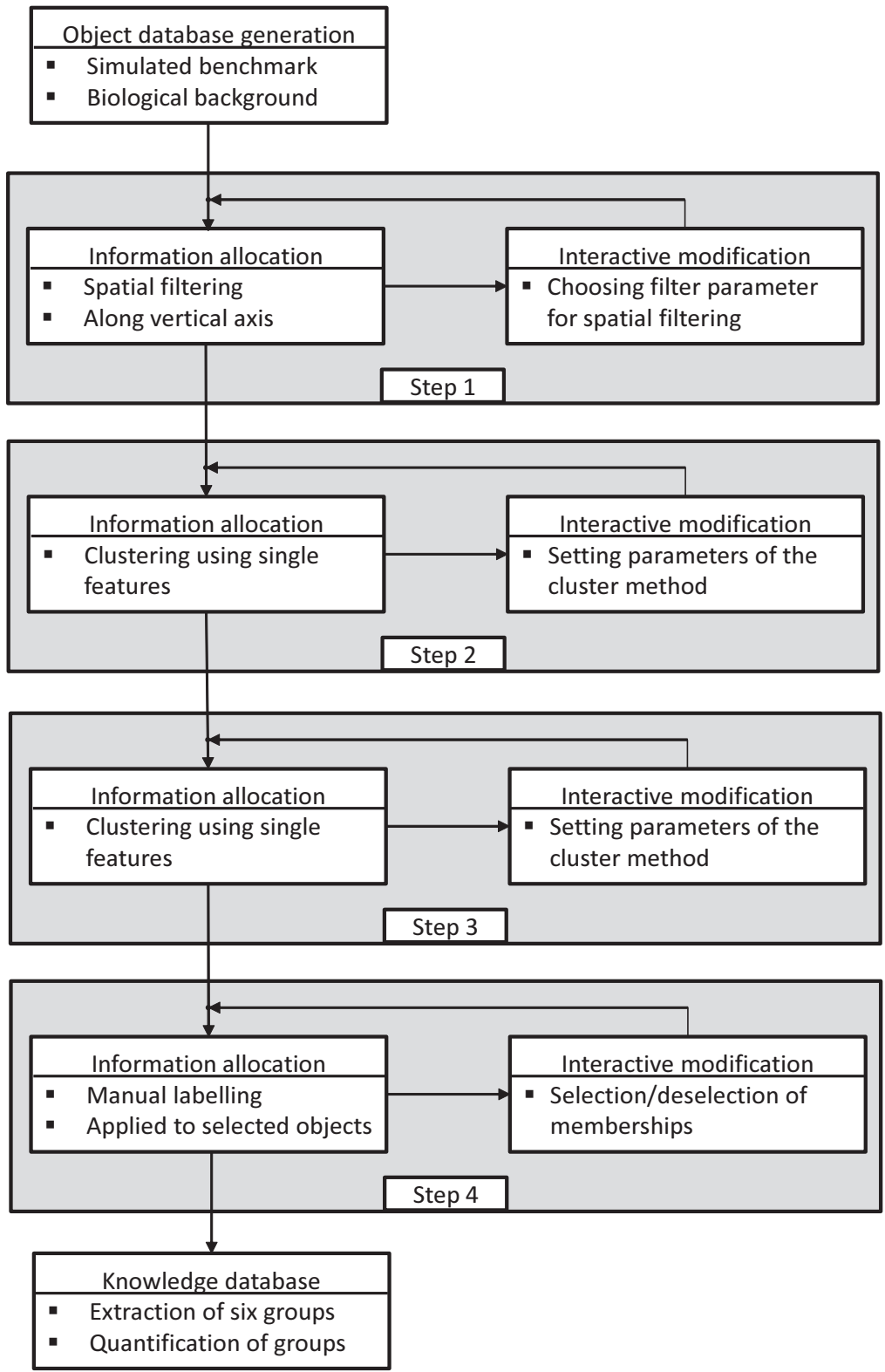

Figure 4: Stepwise application of the KD-process. The "information allocation" and "interactive modification" processes for each step is listed.

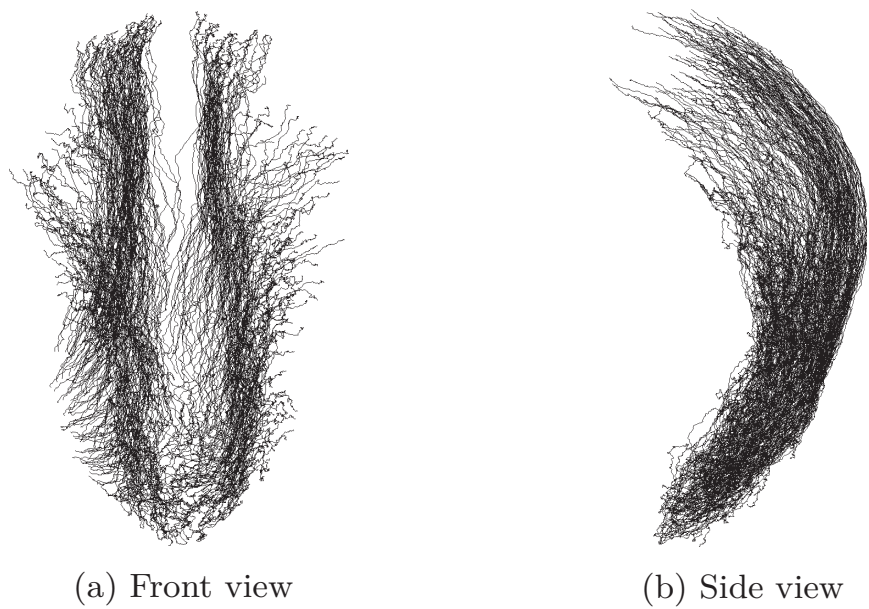

(a) Front view        (b) Side view

Figure 5: The generated Benchmark database. Here the trajectories of moving objects are displayed. The method from [31] is used.

**Overall Problem Formulation**

The overall artificial problem formulation to serve as a basis for the knowledge discovery framework is derived from current real world biological issues. The problem formulation comprised the discovery of six different groups of objects according their qualitatively known spatial occurrence and movement direction in a database containing also many other objects. Here three groups on each side (left and right according the vertical axis in Figure 5) of the benchmark have to be extracted. The step-wise problem formulation to extract these groups is described in more detail. Based on the overall problem formulation the aim of the knowledge discovery framework is to extract the corresponding knowledge database.

**Design of the Knowledge Discovery Process**

The overall problem formulation is integrated step by step within the design of the KD-process to extract the interesting groups of objects. Through this step-wise approach the *a priori* knowledge can be integrated efficiently. In the following the single steps of the iterative design of the KD-process are described in detail. Thereby, the structure within each step contains the verbal problem formulation and the transformation to the formal one, as well as the description of the information allocation process and the interactive modifications that are applied. Finally at the end of each step, the results are highlighted visually.

Step 1 - Interactive filtering: All the objects that are of interest, are located spatially in the upper half according the vertical axis in Figure 5. This verbal problem formulation is transformed to the formal problem formulation task of spatial filtering using a threshold applied to the values of the y-axis (equal the vertical axis) of the data.

For the information allocation process, interactive filtering with a threshold is used. This threshold can be interactively modified by the user. Through visual feedback of the filter result, the user can modify the threshold in the way, that the visualization of the objects agrees with his *a priori* knowledge. The result of interactive filtering is shown in Figure 6.

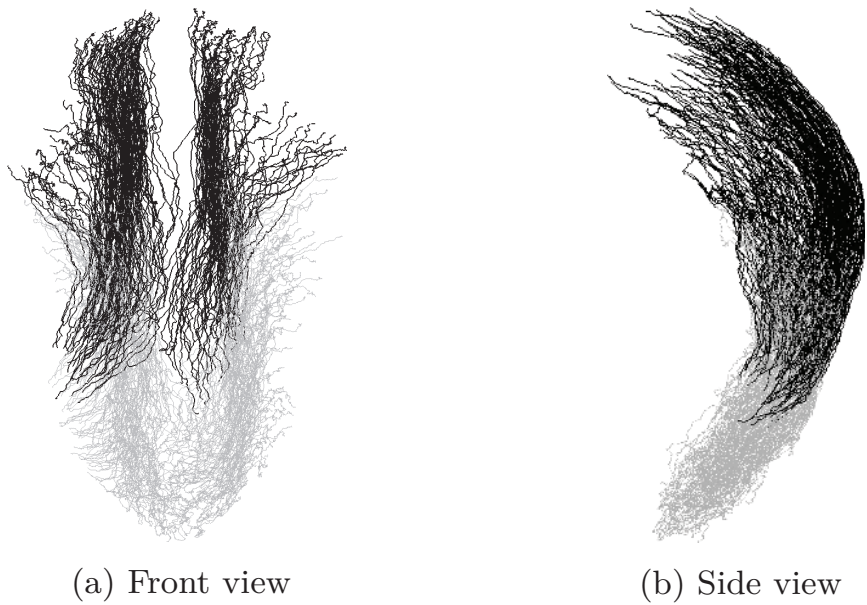

(a) Front view       (b) Side view

Figure 6: Benchmark database: Interactive spatial filtering according the vertical axes. For further analysis the black printed object trajectories are used.

Step 2 - Separation into two sides: The verbal problem formulation is to separate the objects in two halves (left and right assuming a symmetric object structure) in the horizontal direction. The transformed formal problem formulation is to extract the start and end point of each object to separate them. Based on these characteristics a fuzzy c-means clustering using the start points of the objects as single features (x-,y- and z-component) with two cluster centers is applied to the object data (black highlighted in Figure 6). The settings of the cluster method can be used for further interactive modifications. For further steps within the KD-process the two separated halves are considered as a unit (the information of objects belonging to each half is stored for a later splitting of the extracted groups), however the information of the objects belonging to the both sides is retained. The separation into the both halves is shown in Figure 7.

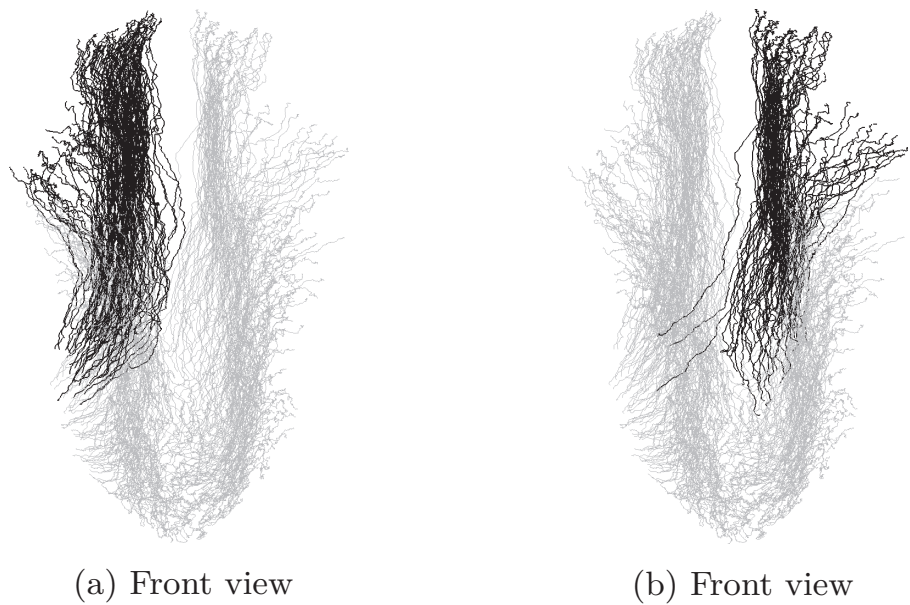

(a) Front view  (b) Front view

Figure 7: Benchmark database: Separation according the sides using clustering: (a) left side; (b) right side

Step 3 - Cluster-based separation: Based on the objects of interest (black highlighted objects in Figure 6) the user wants to further distinguish between two groups. Described in a verbal problem formulation, the first group contain objects having a more straight movement in combination with a location near the "surface". The objects not following these characteristics belong to the second group. The verbal task of having "more straight movement" is transformed to the formal problem formulation (the term "problem formulation" deviated from the original definition in [32]) of calculating the angle of each object trajectory to a reference plane (in this case orthogonal to the front view in Figure 8). Further, the verbal condition of a "location near the surface" is transformed into calculating a single feature that is the range of motion of the objects rotated projection (the whole benchmark is rotated according a predefined angle) on the plane orthogonal to the front view in Figure 8. These two features extracted in the formal problem formulation are further used for clustering. Again a fuzzy c-means cluster method is used with two cluster centers (there are two groups to distinguish). Further the cluster parameters can be used for interactive modifications. Figure 8 shows the group having a more straight movement and is located near the surface (right hand side in the side view). The other group is shown in Figure 9.

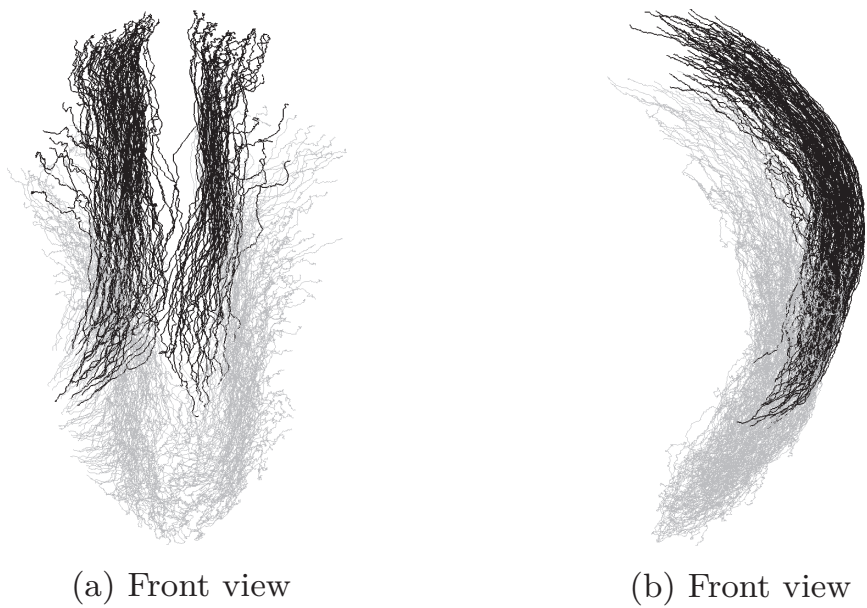

(a) Front view  (b) Front view

Figure 8: Benchmark database: Separating objects having a straight movement according the vertical axes and are further placed on the "surface" (highlighted in black) against all other objects (colored in grey).

Step 4 - Manual information allocation: The second group, extracted in Step 3 (black highlighted in Figure 9), has to be further separated into two groups. The first group exhibits rather forward directed movements and the second group exhibits rather outward directed movements. Through the complex movement patterns of the objects, the ability is used to allocate the information manually to the objects through labeling. Here, the user's *a priori* knowledge is used directly for information allocation. For an effective and fast labeling a visual framework was generated supporting the user within the process of information allocation by clicking to each object for a selection or deselection to a group. The results of manual information allocation are displayed in Figure 10.

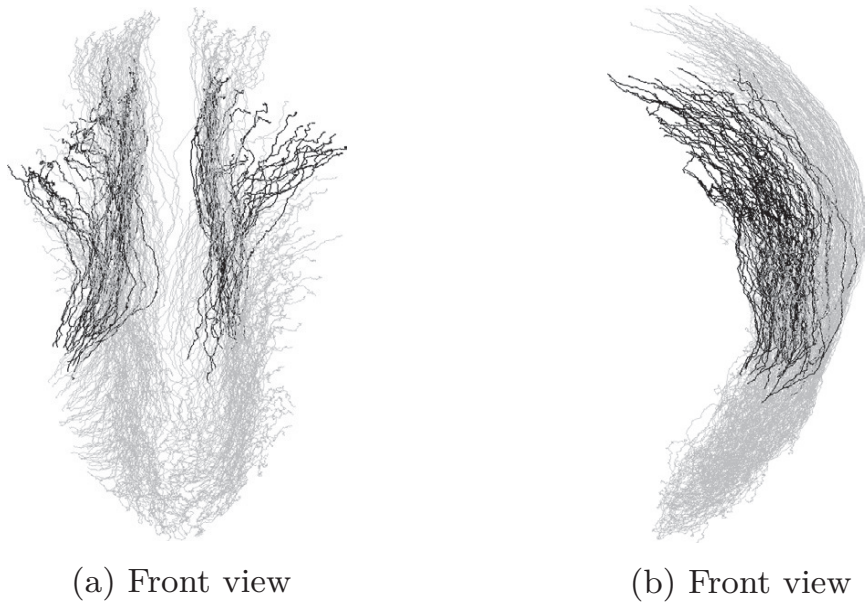

(a) Front view  (b) Front view

Figure 9: Benchmark database: The objects that do not have a straight movement according the vertical axes and additionally are not placed on the "surface".

**Result of the Knowledge Discovery Process**

The result of the design of the knowledge discovery framework applied to the benchmark database based on the problem formulation is shown in Figure 12. In this knowledge database the six extracted groups are visualized, each in front and side view. Here Group 1 and 4 can observed in Figure 10a (left and right side), Group 2 and 3 in Figure 10b (left and right side) and Group 5 and 6 can be observed in Figure 8 (left and right side). Here the heterogeneous spatial distribution of the objects can be shown. Based on the knowledge database further analysis and evaluation steps can be applied. Figure 11 shows one example of the evaluation of the extracted groups according to one chosen characteristic (here the mean curvilinear speed). Similar to this procedure many other characteristics can be investigated and reveal further insights in the object data. This serve as a basis for application-dependent generation of new knowledge about the behavior and characteristics of objects.

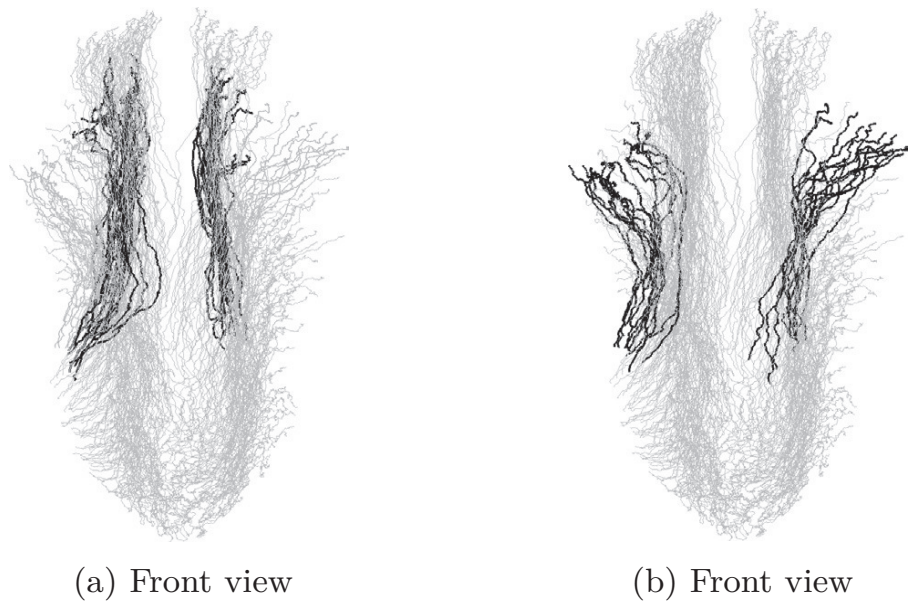

(a) Front view           (b) Front view

Figure 10: Benchmark database: Result of manual information allocation applied to the black highlighted group in Figure 9. Group of objects having a more straight movement direction (a) and the other group (b).

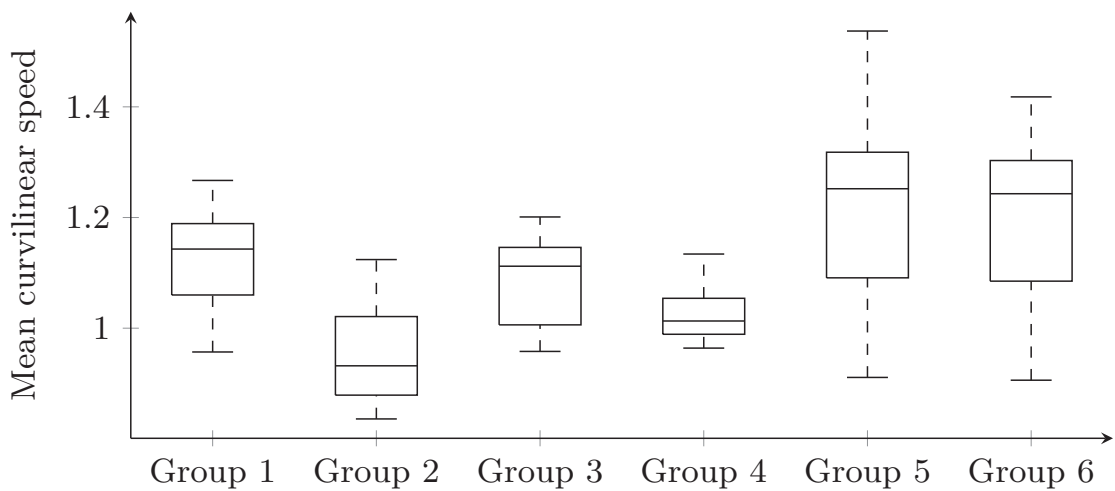

Figure 11: Evaluation of the characteristics "Mean curvilinear speed" (the mean speed from one object position to the following object position) in all groups.

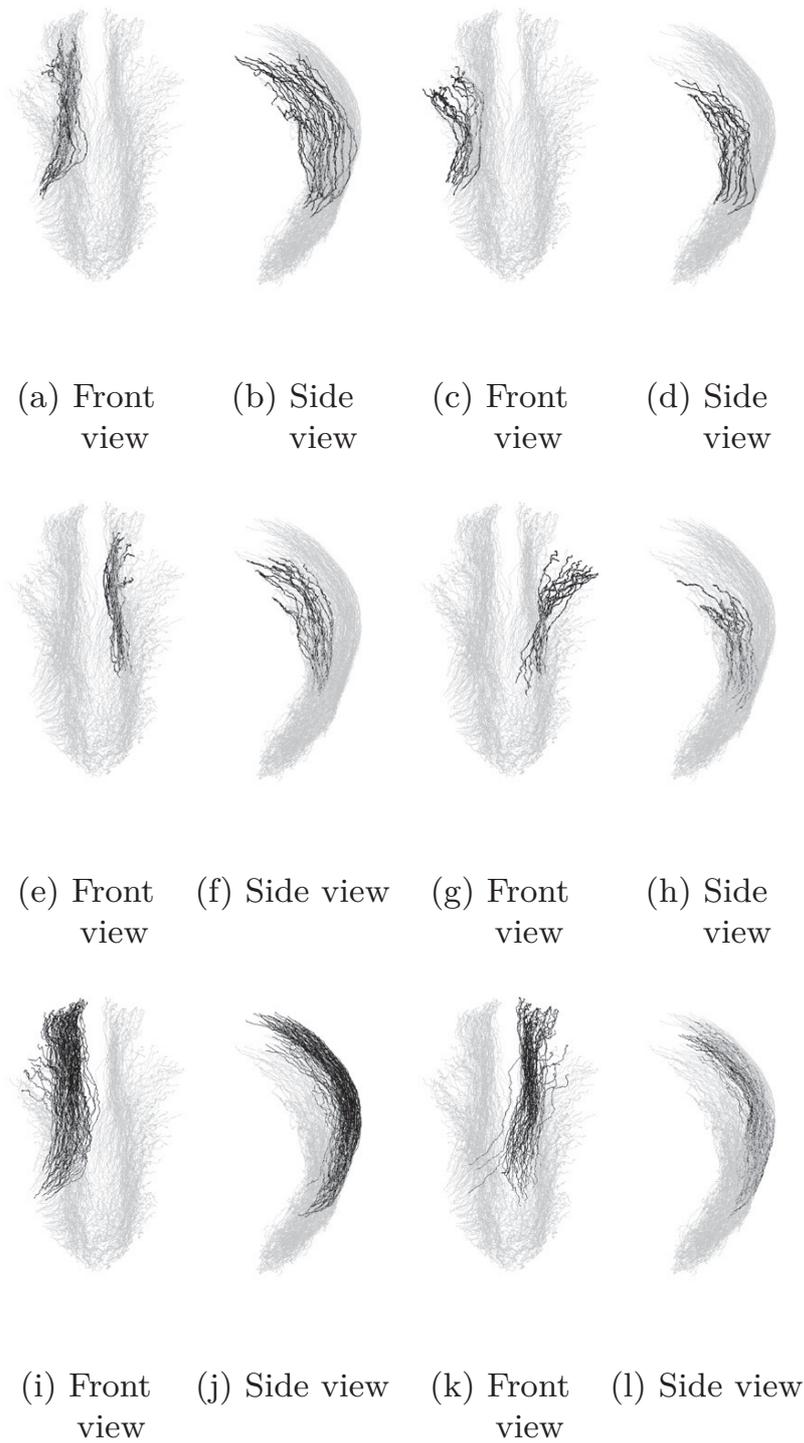

Figure 12: Knowledge database: Extracted groups. Group 1 (a,b), Group 2 (c,d), Group 3 (e,f), Group 4 (g,h), Group 5 (i,j) and Group 6 (k,l).

## Conclusion

Using the knowledge discovery framework, it is possible to explore object databases and extract groups of objects with highly heterogeneous movement behavior by efficiently integrating *a priori* knowledge through interacting with the framework. The whole process is modular expandable and is therefore adaptive to any problem formulation. Further, the flexible use of different information allocation processes reveal a great potential to efficiently incorporate the *a priori* knowledge of different users in different ways.

Therefore, the stepwise knowledge discovery process embedded in the knowledge discovery framework is described in detail to point out the flexibility of such a system incorporating object databases from different applications. The described framework can be used to gain knowledge out of object databases in many different fields. This knowledge can be used to gain further insights and improve the understanding of underlying phenomena. The functionality of the proposed framework is exemplarily demonstrated using a benchmark database based on real biological object data.

A further step will be the extension of the modular framework to handle the occurrence of less quantitative and qualitative *a priori* knowledge. Future work will therefore be put on developing new methods for the information allocation process (e.g. create application-dependent standard feature sets or handle incomplete tracks) and to improve the interactive modifications (application-dependent visual frameworks) within the knowledge discovery process. Additionally, the whole framework can be transferred to other real object databases from several fields to extract valuable knowledge out of the underlying objects to get closer insight and increase the understanding of the object database.